\pdfoutput=1
\documentclass[11pt]{article}

\usepackage[final]{mtsummit25}
\usepackage{copyright}

\usepackage{times}
\usepackage{latexsym}

\usepackage[T1]{fontenc}
\usepackage[utf8]{inputenc}

\usepackage[normalem]{ulem} % sout, uline
\usepackage{xcolor}
\usepackage{microtype}

\usepackage{inconsolata}

\usepackage{graphicx}

\usepackage{booktabs}  % For nice horizontal rules
\usepackage{siunitx}   % For number alignment (optional
\usepackage[capitalize,noabbrev]{cleveref}
\usepackage{xspace}

\usepackage{multirow}
\usepackage{tabularx}
\usepackage{tikz}
\usetikzlibrary{arrows.meta}

\newcommand{\labse}{\texttt{LaBSE}\xspace}
\newcommand{\random}{\texttt{random BERT}\xspace}
\newcommand{\czert}{\texttt{Czert}\xspace}
\newcommand{\simcse}{\texttt{SimCSE}\xspace}
\newcommand{\fernet}{\texttt{FERNET}\xspace}
\newcommand{\xlmr}{\texttt{XLM-R}\xspace}
\newcommand{\robeczech}{\texttt{RobeCzech}\xspace}
\newcommand{\retromae}{\texttt{RetroMAE}\xspace}
\newcommand{\mepet}{\texttt{mE5}\xspace}
\newcommand{\cometWithReference}{\texttt{wmt22-comet-da}\xspace}
\newcommand{\cometReferenceless}{\texttt{wmt22-cometkiwi-da}\xspace}
\newcommand{\meteor}{\textsc{Meteor-NEXT}\xspace}

\newcommand{\cometm}[1]{$\mathrm{CE}_\mathrm{MTE}$(#1)}
\newcommand{\cometq}[1]{$\mathrm{CE}_\mathrm{QE}$(#1)}
\newcommand{\footurl}[1]{\footnote{\url{#1}}}

\title{Intrinsic vs. Extrinsic Evaluation of Czech Sentence Embeddings: \\ Semantic Relevance Doesn't Help with MT Evaluation}

\author{Petra Barančíková \and Ondřej Bojar  \\
 Charles University \\
 Faculty of Mathematics and Physics\\
 Institute of Formal and Applied Linguistics\\
 \texttt{\{barancikova,bojar\}@ufal.mff.cuni.cz} }

\begin{document}
\maketitle
\begin{abstract}
In this paper, we compare Czech-specific and multilingual sentence embedding models through intrinsic and extrinsic evaluation paradigms. For intrinsic evaluation, we employ Costra, a complex sentence transformation dataset, and several Semantic Textual Similarity (STS) benchmarks to assess the ability of the embeddings to capture linguistic phenomena such as semantic similarity, temporal aspects, and stylistic variations. In the extrinsic evaluation, we fine-tune each embedding model using COMET-based metrics for machine translation evaluation.

Our experiments reveal an interesting disconnect: models that excel in intrinsic semantic similarity tests do not consistently yield superior performance on downstream translation evaluation tasks. Conversely, models with seemingly over-smoothed embedding spaces can, through fine-tuning, achieve excellent results. These findings highlight the complex relationship between semantic property probes and downstream task, emphasizing the need for more research into ``operationalizable semantics'' in sentence embeddings, or more in-depth downstream tasks datasets (here translation evaluation).
\end{abstract}

\section{Introduction}

Machine translation (MT) evaluation has advanced significantly in recent years, finally moving beyond traditional surface-level metrics like BLEU \citep{papineni-etal-2002-bleu} towards more sophisticated approaches based on neural networks and contextualized embeddings. 

State-of-the-art MT evaluation metrics such as COMET \citep{rei-etal-2022-cometkiwi} and BLEURT \citep{sellam-etal-2020-bleurt} use sentence embeddings to better capture semantic similarity between translations and references, achieving much higher correlation with human judgments than traditional metrics. 

However, the rapid development of new embedding models presents MT researchers with  a~challenging choice. Although multilingual models such as LaBSE \citep{feng-etal-2022-language} and XLM-RoBERTa \citep{xlm-r} have shown strong cross-lingual capabilities, there are also language-specific models that claim superior performance for a selected language. For morphologically rich languages like Czech, it remains unclear whether these specialized sentence embeddings offer advantages over multilingual alternatives when used in MT evaluation. 

In this paper,  we examine the evaluation of English-to-Czech machine translation and compare several state-of-the-art Czech-specific models against multilingual models using both intrinsic evaluation and extrinsic evaluation. To this end, we see the task of machine translation evaluation (MTE) and quality estimation (QE), i.e. MTE without professionally translated reference sentences, as methods for extrinsic evaluation of sentence embeddings. For intrinsic evaluation, we assess how well the examined sentence embeddings reflect semantic properties exemplified in two datasets: Costra \citep{costra} and Semantic Textual Similarity (STS, \citealp{bednavr2024some}). In sum, our goal is to understand whether the performance of a model in intrinsic semantic tasks correlates with its usability for MT evaluation, potentially simplifying the selection of embeddings.

\section{Related Work}
\label{rw}
Several studies have raised concerns about the use of STS as an evaluation metric. For instance, \citet{reimers-etal-2016-task}, \citet{pitfalls2019}, and \citet{zhelezniak-etal-2019-correlation} argue that, while STS can capture certain semantic similarities, it does not reliably predict how effective sentence representations will be for downstream tasks. These works highlight how STS tasks often encourage surface-level heuristics or oversimplified semantic similarity patterns that may not generalize to more complex applications like entailment or paraphrasing detection. 

To address these limitations, new intrinsic evaluation methods such as EvalRank \citep{wang-etal-2022-just} and SentBench \citep{xiaoming-etal-2023-sentbench} have been proposed, both of which exhibit a~stronger correlation with extrinsic evaluation measures. These benchmarks evaluate sentence representations through information retrieval, sentence ordering, and probing tasks, offering a more holistic view of embedding quality that aligns better with actual downstream task performance. %TODO: jen pro anglictinu?

It is important to note that these previous experiments did not specifically focus on machine translation evaluation, which seems to be very close to~STS---it also involves comparing pairs of sentences to assess their semantic closeness. \citet{cifka-bojar-2018-bleu} report a negative correlation between the translation quality of Transformer models measured by BLEU and the semantic properties (assessed using STS) of the sentence embeddings derived from the Transformer model. In contrast, \citet{libovicky-probing} demonstrate a strong positive correlation between STS performance and translation quality for both Transformer and RNN-based models. %These conflicting results underscore the complexity of linking intrinsic evaluation metrics to performance in machine translation tasks.

More recently, \citet{freitag-etal-2022-results} have advocated for the use of semantic-aware metrics such as BERTScore \citep{bertscore} and COMET \citep{rei-etal-2020-comet} in MT evaluation, showing that these outperform BLEU in correlating with human judgments. These models incorporate contextual embeddings and often exhibit closer alignment with human-perceived meaning, bringing MT evaluation closer to the goals of intrinsic semantic understanding.

\section{Models}
For our experiments, we used several state-of-the-art sentence embedding models, employing both Czech-specific and multilingual variants. The Czech encoders include three \emph{base}-size Transformer architectures, each using masked language modeling as their primary pretraining objective---CZERT-b-cased  (\czert, \citealp{sido2021czert}), FERNET-C5 (\fernet, \citealp{FERNETC5}) and \robeczech \citep{robeczech}.

To provide a broader comparative analysis, we also experimented with multilingual sentence embedding models trained on datasets that contain Czech texts. These include \labse \citep{feng-etal-2022-language}, a model generating language-agnostic representations for more than one hundred languages with remarkable cross-lingual alignment, since its training objective was machine translation. 
Furthermore, we evaluated two \emph{large} models: XLM-RoBERTa-large (\xlmr, \citealp{xlm-r}) and multilingual-e5-large (\mepet, \citealp{wang2024multilingual}), a model pretrained using a contrastive learning approach on a diverse range of tasks, including natural language inference and question answering across multiple languages.

As a baseline model, we employed a BERT architecture \citep{bert} with randomly initialized weights.
%, using Xavier normal initialization. 
The only component inherited from the pretrained \emph{`bert-base-multilingual-cased'} model is the tokenizer. This means that while the model processes input according to the tokenization patterns learned from multilingual data, it does not benefit from any pretrained language representations. We refer to this configuration as \random. This setup isolates and assess the contribution of the tokenizer alone, establishing a lower performance bound and offering a meaningful point of comparison to evaluate the benefits of pretraining.

\section{Intrinsic Evaluation}
We first evaluate the embeddings using a series of semantic benchmarks to determine their ability to accurately capture various semantic properties of a sentence.

\subsection{Costra}
As the first dataset for intrinsic evaluation, we used the Costra\footurl{https://github.com/barancik/costra} dataset \citep{costra}. It was created manually, specifically to test the quality of Czech embeddings, focusing on complex transformations of sentences beyond standard paraphrasing or simple word-level changes. The sentence embeddings are tested across the following six categories: 

\begin{itemize}
    \item \textbf{Basic}: evaluates whether paraphrases are positioned closer together in embedding space compared to transformations that significantly alter the meaning of the original sentence. 
    \item \textbf{Modality}: measures whether paraphrases are more similar to their original sentence than transformations that change the sentence's modality (e.g., possibility or prohibition).
    \item \textbf{Time, Style, Generalization, Opposite}: these categories test embeddings' ability to reflect linear ordering of sentence variations (e.g., from the least general to the most general) as proposed by annotators.
\end{itemize}

Each category is scored on a scale from 0 (worst) to 1 (best), reflecting the proportion of Costra sentence triplets for which the relations in the sentence vector space align with human annotations. For example, consider a triplet consisting of a seed sentence $S$, its paraphrase $P$, and its opposite sentence $O$. Ideally, the cosine similarity $S_{C}$ should satisfy $  S_{C}(S,P) >  S_{C}(S,O) $ and $S_{C}(S,P) >  S_{C}(P,O) $, indicating that the model correctly identifies the paraphrase closer to the seed sentence than the opposite sentence.

\begin{table*}
  \begin{tikzpicture}
  \node (table) {
     \begin{tabular}{l|r|rrrrrr|rr}
        \multicolumn{4}{c}{} & \multicolumn{4}{c}{\tikz \draw[<->, thick] (0,0) -- node[above] {Costra--} (5,0);} \\
        \multicolumn{2}{c}{} & \multicolumn{6}{c}{\tikz \draw[<->, thick] (0,0) -- node[above] {Costra} (7,0);} \\

    \hline
    \textbf{Embeddings} & \textbf{Size} & \textbf{Basic} & \textbf{Mod.} & \textbf{Time} & \textbf{Style} & \textbf{Gen.} & \textbf{Opp.} & \textbf{Costra}  & \textbf{Costra--}\\
    \hline
\simcse    & 256 & 0.20 & 0.35 & \textbf{0.74} & 0.63 & 0.73 & \textbf{0.78} & \textbf{0.57} & \textbf{0.72} \\
\mepet   & 1,024 & 0.24 & 0.34 & 0.71 & 0.62 & \textbf{0.75} & 0.77 & \textbf{0.57} & 0.71 \\
\labse     & 768 & 0.20 & 0.26 & 0.71 & 0.63 &\textbf{ 0.75} & 0.75 & 0.55 & 0.71 \\
\retromae  & 256 & 0.06 & 0.06 & 0.69 & 0.63 & 0.70 & 0.76 & 0.48 & 0.70 \\
\robeczech & 768 & 0.15 & 0.13 & 0.69 & \textbf{0.65} & 0.69 & 0.75 & 0.51 & 0.70 \\
\random    & 768 & 0.08 & 0.06 & 0.65 & 0.60 & 0.72 & 0.73 & 0.47 & 0.68 \\
\czert     & 768 & 0.31 & 0.35 & 0.66 & 0.64 & 0.69 & 0.69 & 0.56 & 0.67 \\
\xlmr    & 1,024 & 0.16 & 0.11 & 0.65 & 0.61 & 0.67 & 0.68 & 0.48 & 0.65 \\
\fernet    & 768 & 0.33 & 0.38 & 0.65 & 0.61 & 0.63 & 0.68 & 0.54 & 0.64 \\
random vectors & 256 &\textbf{0.50} & \textbf{0.51} & 0.49 & 0.50 & 0.49 & 0.50 &  0.50 & 0.50 \\
\end{tabular}
  \label{tab:costra2}
    };
    \end{tikzpicture}
      \caption{This Table presents the results of intrinsic evaluation using the Costra dataset. The Costra score ranges from 0 (worst) to 1 (best) in each category. The overall Costra score is calculated as the arithmetic mean across all categories. Costra-- represents the mean score excluding the first two categories (Basic and Mod.), as these categories appear excessively challenging for all pretrained encoders evaluated.
      }
  \label{tab:costra}
\end{table*}

The results are presented in \Cref{tab:costra}, with the overall Costra score calculated as the arithmetic mean across all six categories.  In particular, the evaluation shows that all sophisticated models failed to outperform randomly generated embeddings\footnote{Not to be confused with \random, we evaluated Costra also using completely \emph{random vectors}.} in the first two categories, \textbf{Basic} and \textbf{Modality}. In fact, these categories were designed to be particularly challenging, including comparisons of paraphrases with substantial lexical variation and sentences that, despite the close lexical similarity to a paraphrase, differ significantly in meaning. These results suggest that all models were fooled by surface-level similarity, making randomly generated embeddings the overall winner in these two categories. Consequently, it is impossible to distinguish whether slight improvements in these categories can be attributed to model quality or to randomness.

To address this limitation, we introduce the \textbf{Costra--} score, calculated as the average of the four remaining categories: \textbf{Time}, \textbf{Style}, \textbf{Generalization}, and \textbf{Opposite}. However, the \textbf{Costra--} scores revealed only marginal differences across models. The smallest model, \simcse, slightly outperformed its counterparts but the improvement was not substantial. In fact, the models performed only marginally better than the \random model, suggesting limited success in capturing phenomena tested in the Costra dataset, such as linearity of time or generalization. Several models, including large \xlmr, even performed worse than \random.

\begin{table}[h!]
    \centering
 \begin{tabular}{l|c}
 \hline
\textbf{Embeddings} & \textbf{avg. similarity}\\
\hline 
\simcse & \textbf{87.83} \\
\labse & 82.91 \\
\mepet & 78.39 \\
\retromae & 76.30 \\
\czert & 74.79 \\
\robeczech & 70.28 \\
\fernet & 65.46 \\
\random & 60.48 \\
\xlmr & 57.88 \\
\end{tabular}
    \caption{Results of intrinsic evaluation on three STS datasets.}
    \label{tab:STS}
\end{table}

\subsection{Semantic Textual Similarity}
\Cref{tab:STS} presents the results of our evaluation of sentence embeddings on the Semantic Textual Similarity (STS) task. Performance is measured using the automated evaluation tool\footurl{https://github.com/seznam/czech-semantic-embedding-models} provided by \citet{bednavr2024some}. This tool computes similarity for pairs of sentences in three STS datasets. For precomputed sentence embeddings, it explores different embedding similarity metrics including cosine similarity, dot product, and Manhattan distance. Additionally, it applies various sentence embeddings pooling strategies and selects the highest average score as the final result.

Interestingly, the results are consistent with findings from Costra, with \simcse being the overall best performing model, followed by \mepet and \labse in the next two positions. Surprisingly, \xlmr, despite being a powerful multilingual model, may not be well-optimized for Czech-specific STS tasks, ranking last in the evaluation, performing even worse than \random.

\section{Extrinsic Evaluation---MTE and QE}

Extrinsic evaluation utilizes sentence embeddings as feature vectors for machine learning algorithms in downstream NLP tasks---MTE and QE in our tasks. It serves well to choose the best method for a particular task but not as an absolute metric of embedding quality, as the performance of the embeddings does not correlate across different tasks \citep{bakarov}. 
%TODO: toto chce upravit, nebo hodit do RW, bakarov mluvi o vetnych embeddingach

\subsection{Data}
In the following experiments, we utilize datasets from the Workshop on Machine Translation (WMT), selecting data from English-to-Czech translations. These datasets include English source sentences, Czech hypotheses (i.e., machine translated outputs), Czech reference sentences, and the human translation quality scores collected using the Direct Assessment (DA) method \citep{graham-etal-2013-continuous} and subsequently \texttt{z}-normalized.

Data from WMT17 to WMT19 \citep{wmt2017, bojar-etal-2018-findings, barrault-etal-2019-findings} were used to train the COMET estimators \citep{rei-etal-2020-comet}. The validation of the models was performed on the WMT20 dataset \citep{barrault-etal-2020-findings}, and the performance of the models was tested using the WTM21 \cite{akhbardeh-etal-2021-findings} and WMT22 \citep{kocmi-etal-2022-findings} datasets.

\subsection{MTE Baseline Approach}
\begin{table*}[]
    \centering
\begin{tabular}{l|ccc|ccc|c}
\hline
 \multicolumn{1}{c|}{\textbf{Sentence}} &  \multicolumn{3}{c|}{\textbf{WMT21 test set}} &  \multicolumn{4}{c}{\textbf{WMT22 test set}} \\
\textbf{Embeddings} & $\rho_{H,R}$ & $\rho_{H,S}$ & $S_{C}(S,R)$ & $\rho_{H,R}$  & $\rho_{H,S}$ & $S_{C}(S,R)$  & $S_{C}(S_{R},R_{R})$ \\
\hline
\mepet     & \textbf{0.29} & 0.04 & 0.89 & 0.26 & 0.01 & 0.90 & 0.75 \\
\retromae  & 0.26 & -0.10 & 0.76 & \textbf{0.27} & \textbf{0.09} & 0.76 & 0.69 \\
\simcse    & 0.24 & \textbf{0.13} & 0.85 & 0.25 & 0.05 & 0.82 & 0.09 \\
\czert     & 0.20 & -0.03 & 0.63 & 0.18 & -0.06 & 0.62 & 0.52 \\
\xlmr      & 0.17 & -0.08 & 1.00 & 0.05 & -0.10 & 1.00 & 0.99 \\
\robeczech & 0.15 & -0.16 & 0.92 & 0.11 & -0.06 & 0.91 & 0.89 \\
\labse     & 0.11 & 0.03  & 0.89 & 0.19 &  0.06 & 0.88 & 0.31 \\
\fernet    & 0.07 & -0.11 & 0.45 & 0.11 & -0.03 & 0.40 & 0.35 \\
\random    & 0.06 & -0.16 & 0.99 & 0.03 & -0.20 & 0.98 & 0.98 \\
\end{tabular}
    \caption{Results for baseline MTE approach---using sentence embeddings for direct evaluation without fine-tuning. $\rho_{H,R}$ represents Pearson correlation between human quality assessments and the cosine similarity between the translation hypothesis and the reference translation, while $\rho_{H,S}$ shows the correlation between human judgments and the cosine similarity between the hypothesis and the source sentence. $S_{C}(S,R)$ represents cosine similarity between references and sources. The last column represents cosine similarity between randomly shuffled source and reference sentences averaged over 100 runs.}
    \label{tab:baseline}
\end{table*}

Before fine-tuning the sentence embedding models for machine translation evaluation, we conducted a preliminary analysis to assess their default ability to evaluate translation quality. Specifically, we examined Pearson's correlation between human judgments and the cosine similarities computed between (i) a hypothesis and a reference translation and (ii) a hypothesis and a source sentence. We expected high cosine similarity for multilingual models, reflecting their ability to capture cross-lingual semantic relationships, whereas Czech-specific models---lacking such cross-lingual information---were anticipated to have random similarity scores.

Furthermore, we examined the intrinsic quality of the embedding spaces by measuring the cosine similarity between the source and reference embeddings. We also performed a random shuffling experiment designed to evaluate the discriminative ability of the embeddings.

The results presented in \Cref{tab:baseline} reveal that even without
fine-tuning, a slight correlation between human judgments and cosine similarity
of hypotheses and references is observable in certain models---particularly \mepet, \retromae, and \simcse. However, contrary to expectations, this does not hold for source sentences; no relationship was detected between human evaluation scores and the cosine similarity computed between a translated sentence and its source sentence, even among the multilingual models.

The analysis of the embedding space via similarity between source and reference sentences provides further insights.  In line with our hypothesis, \xlmr exhibits a near perfect similarity between the source and reference sentences, indicative of a tightly clustered or language-agnostic representation;  however, the same holds for \random. 

To further investigate this behavior, we repeated the experiment using random shuffle of source and reference sentences; see the last column of \Cref{tab:baseline}. The similarity remained perfect for both \xlmr and \random even on shuffled pairs, indicating an overly invariant embedding space, where even pairs of semantically unrelated sentences tend to cluster together. This \emph{over-smoothing} reduces the model's capacity to distinguish subtle differences that are essential for evaluating translation quality. In such cases, even bad translations can receive high similarity scores, lowering the correlation with human judgment. This also explains the poor performance of \xlmr in our intrinsic evaluation task, especially in STS (\Cref{tab:STS}).
More broadly speaking, it casts doubts on any results based on the direct similarity of \xlmr embedding vectors in the Czech language, given that \xlmr assigns similar vectors to random Czech sentences.

\subsection{Models fine-tuning for MTE and QE}

For all sentence encoders, we fine-tuned two COMET-based estimators (CE, \citealp{rei-etal-2020-comet}), one for machine translation evaluation using reference sentences and the other for quality estimation without reference sentences. The COMET models use a dual-encoder architecture: the source sentence, reference translation, and hypothesis are each processed independently using transformer encoder models followed by two hidden layers of sizes 3072 (resp. 2048 for QE) and 1024.

We used the default training settings with the AdamW optimizer ($1.5 \cdot 10^{-5}$ for the regression layers and $1.0 \cdot 10^{-6}$ for the encoder) and a layer-wise decay of 0.95. To preserve encoder generalization, the embeddings were frozen for the first 0.3 epochs. Both models used mixed-layer pooling with a sparsemax-based transformation before pooling and were optimized with mean squared error loss (using a dropout of 0.1). Training was conducted over five epochs, and we selected the checkpoint with the highest Kendall's tau validation on a held-out validation dataset.

These settings were applied consistently across all models without extensive hyperparameter tuning. In total, we trained a total of 18 COMET estimators. To avoid confusion with the original embeddings, we refer to a trained COMET estimator for given embeddings X as to \cometm{X} for machine translation with reference sentences and \cometq{X} for the referenceless quality estimation metric (e.g., for the \czert embeddings, we use \cometm{\czert} and \cometq{\czert}, respectively).

\subsection{Results of MTE and QE evaluation}

\begin{table*}[htb!]
\centering
\begin{tabular}{c|rr|rr}
 & \multicolumn{2}{c}{system-level} & \multicolumn{2}{|c}{segment-level} \\
\hline
\textbf{MTE metrics} & 2021 & 2022 & 2021 & 2022 \\
 \hline
\cometm{\fernet}      &\textbf{ 0.98} & \textbf{0.97} & 0.60 & 0.47 \\
\cometWithReference   & 0.97 & 0.93 & \textbf{0.66} & \textbf{0.51}\\
\cometm{\czert}       & 0.96 & 0.93 & 0.57 & 0.43 \\
\cometm{\xlmr}        & 0.96 & 0.93 & 0.62 & 0.47 \\
\cometm{\robeczech}   & 0.97 & 0.92 & 0.58 & 0.44 \\
\cometm{\mepet}       & 0.96 & 0.92 & 0.59 & 0.46 \\
\meteor               & \textbf{0.98} & 0.84 & 0.24 & 0.21 \\
chrF2                 & 0.97 & 0.84 &   -  & -    \\
\cometm{\retromae}    & 0.91 & 0.82 & 0.43 & 0.34 \\
\cometm{\labse}       & 0.89 & 0.79 & 0.56 & 0.45 \\
\cometm{\simcse}      & 0.96 & 0.74 & 0.44 & 0.37 \\
1-TER                 & 0.95 & 0.60 & -     & - \\
BLEU                  & 0.94 & 0.54 & -     &   -  \\
\cometm{\random}      & 0.35 & -0.35  & 0.23 & 0.22 \\

\multicolumn{5}{c}{} \\
 & \multicolumn{2}{c}{system-level} & \multicolumn{2}{|c}{segment-level} \\
 \hline
\textbf{QE metrics} & 2021 & 2022 & 2021 & 2022 \\
 \hline
\cometq{\fernet}     & \textbf{0.98} & \textbf{0.96}  & 0.60 & 0.46 \\
\cometReferenceless  & 0.95 & 0.91 & \textbf{0.67} & \textbf{0.49} \\
\cometq{\xlmr}       & 0.96 & 0.88 & 0.63 & \textbf{0.49} \\
\cometq{\robeczech}  & 0.97 & 0.87 & 0.58 & 0.39 \\
\cometq{\czert}      & 0.95 & 0.86  & 0.57 & 0.39 \\
\cometq{\mepet}      & 0.93 & 0.76 & 0.59 & 0.45 \\
\cometq{\labse}      & 0.83 & 0.39  & 0.54 & 0.40 \\
\cometq{\retromae}   & 0.64 & 0.15 & 0.39 & 0.23 \\
\cometq{\random}     & 0.47 & -0.19  & 0.26 & 0.20 \\
\cometq{\simcse}     & 0.12 & -0.92  & 0.38 & 0.24 \\
\end{tabular}
\caption{Correlations between human scores and evaluation metrics, including both fine-tuned COMET-based metrics and traditional metrics, computed at the system and segment levels.}
\label{tab:correlations}
\end{table*}

We compare the performance of trained evaluation metrics at the system level with traditional string-matching MTE metrics. Specifically, we include BLEU \citep{papineni-etal-2002-bleu}, TER \citep{snover-etal-2006-study}, and ChrF \citep{popovic-2015-chrf}, using their default configurations as implemented in SacreBLEU \citep{post-2018-call}. % \todo{add sacrebleu signature strings}. 
Additionally, we employ \meteor  \citep{denkowski-lavie-2010-meteor}, a metric that include paraphrase support, on both system and segment levels.

Furthermore, we compute scores using the official pretrained COMET models for machine translation evaluation, namely \cometWithReference \citep{rei-etal-2022-comet}, and for quality estimation, specifically \cometReferenceless \citep{rei-etal-2022-cometkiwi}. These COMET models extend beyond a simple trained COMET estimator, as they incorporate an ensemble approach combining a COMET estimator trained on DA data and sequence predictors trained on MQM annotations.

The results in \Cref{tab:correlations} indicate a clear advantage for COMET-based evaluation metrics over traditional metrics in MTE. In the system-level analysis, the COMET variants \cometm{\fernet} and \cometq{\fernet} achieved consistently remarkably high correlation outperforming even the official COMET ensemble metrics --  \cometWithReference and \cometReferenceless, which were the top-performing metrics at the segment level.

In contrast, classical metrics, although competitive in 2021, showed significant performance degradation in 2022. \cometm{\random} failed completely, highlighting the importance of using pretrained sentence embeddings, even though \cometq{\random} outperformed \cometq{\simcse}, even though \simcse was the best performing encoder in intrinsic evaluation.

Another interesting observation is the small difference in correlations of the top performing embeddings between MTE and QE. The correlation of \cometm{\fernet} and \cometq{\fernet} is practically equal at both the system and segment levels, as if these metrics no longer have use for reference translations. This is consistent with recent research showing that reference-free evaluation has become competitive with reference-based evaluation \citep{rei-etal-2021-references} or even outperforms it \citep{Moosa2024MTRankerRM}.

%\todo{\citet{ma-etal-2019-results} show that metrics struggle on segment level performance despite achieving impressive system level correlation }

\section{Results and Discussion}

When comparing the results of MTE and QE with those of the intrinsic evaluation tasks, we can observe an interesting inversion. Although both evaluation approaches aim to capture semantic similarity, the performance of the embeddings changed significantly after fine-tunings. Specifically, \xlmr and \fernet embeddings, which performed poorly in intrinsic evaluation, became the best performing MTE and QE metrics. In contrast, \simcse, which dominated intrinsic evaluations, ranks among the worst performing metrics in MTE and QE. These results are in line with related research (\Cref{rw}), which shows that STS performance may not accurately predict effectiveness in downstream tasks.

%The most straightforward answer would be a mismatch between the tasks objectives. The intrinsic evaluation tasks measure general semantic similarity, which does not necessarily align with the Direct Assignment scores, which were shown not to be very reliable \citep{kocmi-etal-2024-error}. \todo{i kdyz budu tvrdit, ze DA neni spolehlivy, tak jak to, ze paks ty vysledky jsou konzistentni napric roky i tasky?}

There are several plausible hypotheses that might explain these discrepancies. Let us at least mention them here--- unfortunately, their thorough testing is beyond the scope of this article. 

First, \xlmr and \fernet might perform poorly in intrinsic tasks because their representation spaces are not tuned for fine-grained semantic differences. However, when fine-tuned on a translation quality task, the model might learn to emphasize those aspects of the embedding space that are important for distinguishing translation quality. 

The fine-tuning process for COMET-based evaluation might be effectively reconfiguring the \xlmr embedding space, transforming its initially over-smoothed representations into task-specific features that are highly discriminative for translation quality. Although \xlmr raw embeddings appear to be all clustered together (see \Cref{tab:baseline}), the fine-tuning may introduce transformations that re-weight and separate the dimensions relevant for capturing translation errors. In contrast, \simcse embeddings, which are already optimized for intrinsic semantic discrimination, might leave less room for adjustments necessary to learn the new training objective.

We should also not forget about the different embedding sizes, which played an important role in the observed behavior. The \emph{small} embeddings---\simcse and \retromae---were among the worst performing COMET estimators. Large embeddings, such as those produced by \xlmr, offer a higher-dimensional space that can capture more nuanced semantic and syntactic features. When fine-tuning with the COMET estimator---which adds two hidden layers with sizes 3072 (resp. 2048) and 1024---the richer representation provided by larger embeddings could allow the model to extract and emphasize the translation-specific signals more effectively. 

Interestingly, we can see not too much difference between the monolingual vs. multilingual embedding performances---they seem to be equally represented among the best performing embeddings in both intrinsic and extrinsic tasks. The size of the embeddings seem not to matter in the intrinsic tasks---the top 3 best performing embeddings (\simcse, \labse and \mepet) are \emph{small}, \emph{base} and \emph{large}, respectively. 

\begin{figure*}
    \centering
    \includegraphics[width=0.8\linewidth]{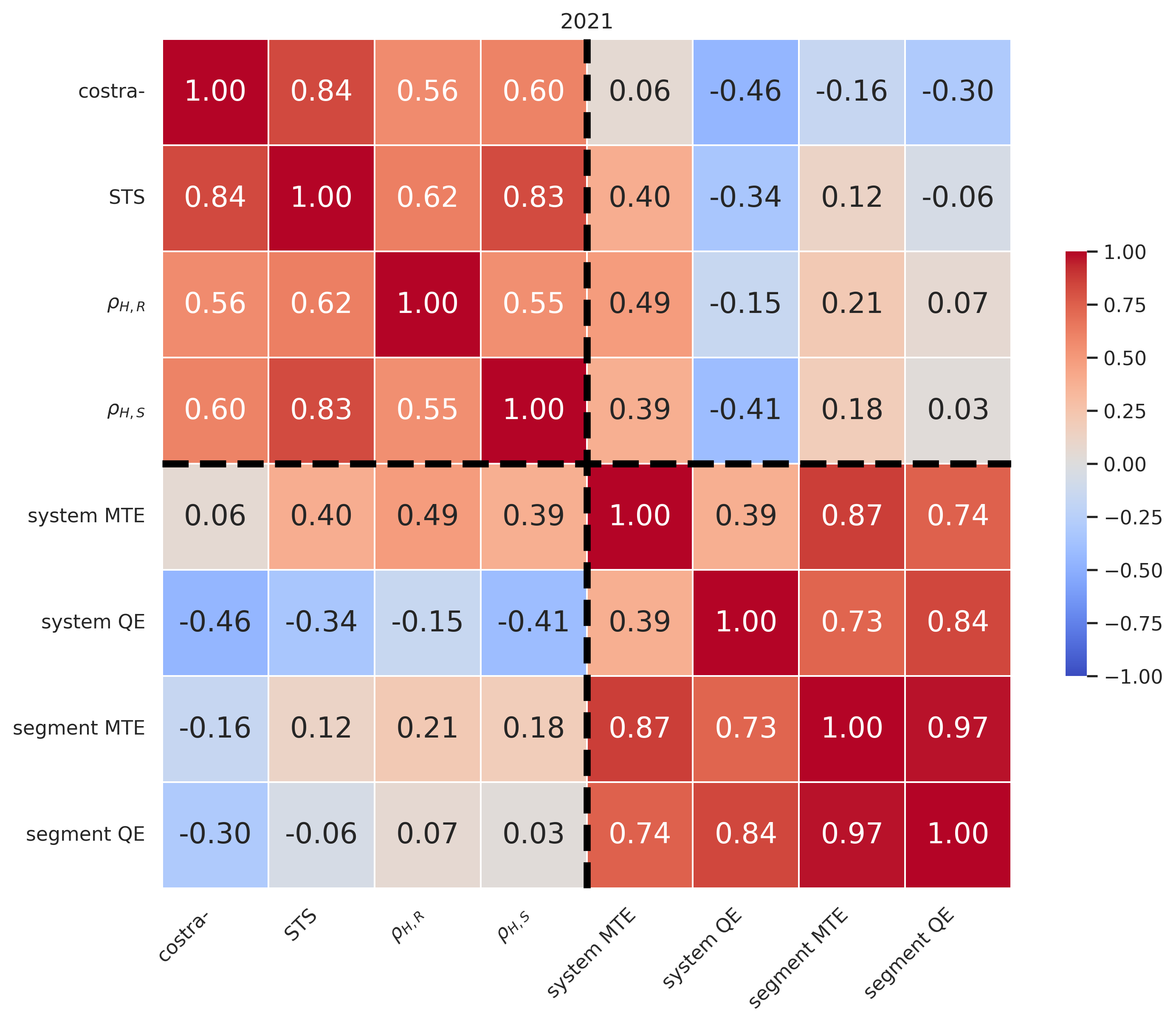}
    \includegraphics[width=0.8\linewidth]{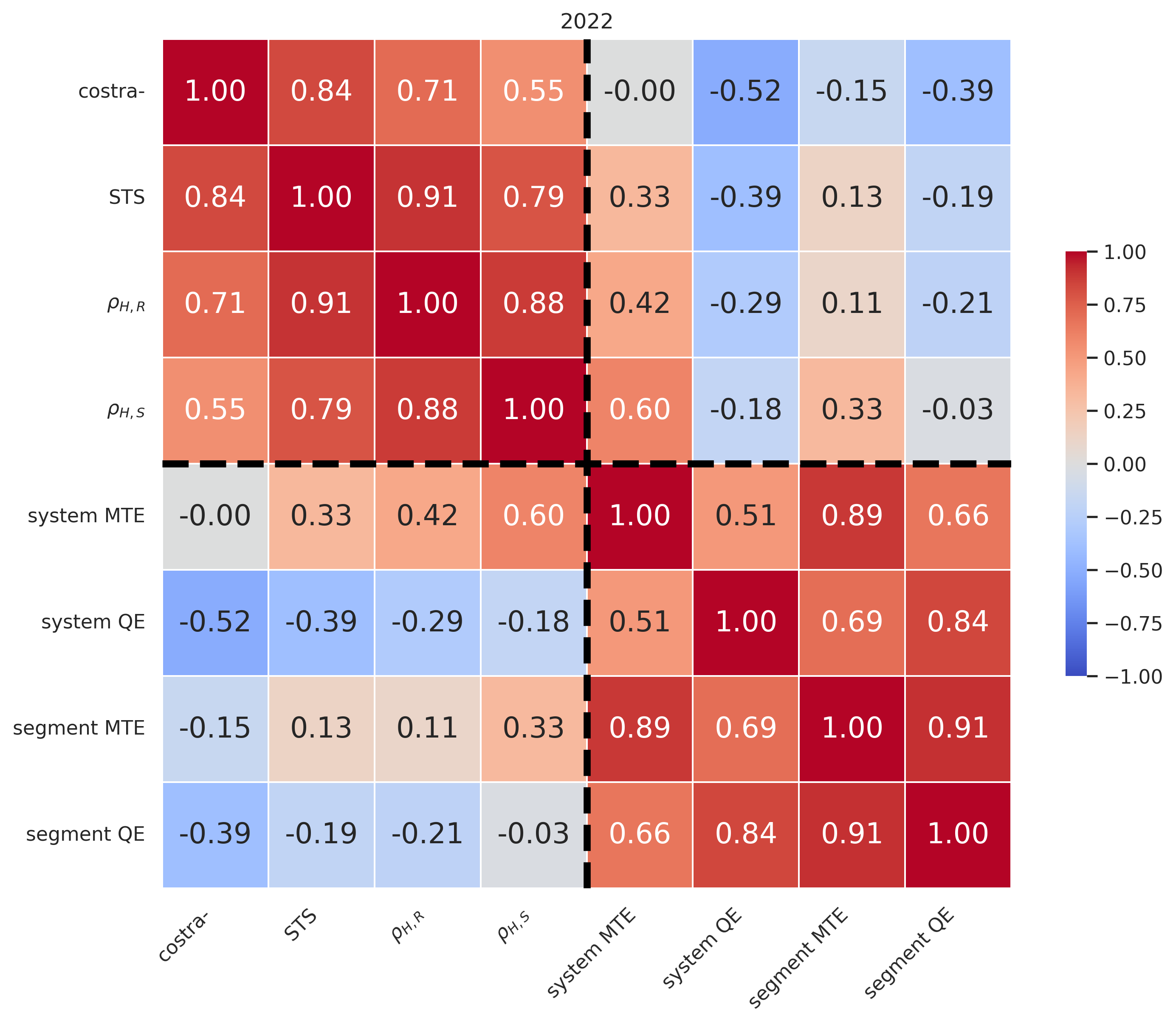}
    \caption{Correlation heatmaps for different method of embedding evaluations. The heatmaps are intentionally kept in a rectangular shape to emphasize the mismatch in correlation patterns between intrinsic evaluation (Costra--, STS, $\rho_{H,R}$, $\rho_{H,S}$) and extrinsic evaluation (system MTE/QE and segment MTE/QE).}
    \label{fig:heatmap}
\end{figure*}

The correlation analysis between different evaluation methodologies, visualized by heatmaps in \Cref{fig:heatmap}, reveals interesting patterns of how different evaluation methodologies relate to each other. These patterns provide valuable insight into the reliability and consistency of various embedding evaluation approaches.

The heatmaps highlight a strong alignment among all intrinsic tasks (Costra--, STS, and MTE baselines). Moreover, there is a strong correlation between the segment-level and the system-level metrics, indicating that aggregated segment scores provide reliable system-level insights. In particular, we observe strong correlations between segment-level metrics (\emph{segment MTE} and \emph{segment QE} showing correlations of 0.97 and 0.91 for 2021 and 2022 respectively), suggesting that these evaluation approaches capture similar aspects of translation quality despite their methodological differences.

%The relationship between reference-based and reference-free evaluation methods also presents interesting patterns. Although system-level MTE and QE show moderate positive correlation (0.39 in 2021, increasing to 0.51 in 2022), their segment-level counterparts demonstrate much stronger agreement (0.97 in 2021, 0.91 in 2022). This suggests that while these methods may diverge in system-level assessment, they tend to agree more strongly when evaluating individual translations.

However, one of the most striking findings is the weak and sometimes even negative correlation between intrinsic evaluation metrics (Costra--, STS) and the system-level quality estimation scores \emph{system QE}. This discrepancy is particularly evident in the 2022 data, where Costra-- shows a negative correlation (-0.52) with \emph{system QE}, challenging the assumption that better semantic representation capabilities necessarily translate to improved MT evaluation performance.

These findings indicate that intrinsic measures, while useful for general semantic similarity, may not sufficiently reflect translation-specific nuances required for MTE or QE. Consequently, intrinsic criteria alone appear inadequate for selecting optimal sentence embeddings for these specific tasks. Further research is needed to identify intrinsic evaluation methods that better capture the subtleties relevant to machine translation. Additionally, it would be valuable to explore in more detail the types of errors penalized in manual MT quality assessments to determine whether these errors predominantly concern sentence meaning or other aspects that should be preserved in translation.

\section{Conclusion and Future Work}
We experimented with several evaluation methods for both Czech and multilingual sentence embeddings, considering intrinsic semantic tasks and downstream application in machine translation evaluation and quality estimation. Our key findings include the following:

\begin{itemize}
    \item \textbf{Intrinsic vs. Extrinsic Discrepancy:} The lack of correspondence between the intrinsic and extrinsic metrics used in our experiments suggests that intrinsic evaluation methods employing these metrics cannot reliably predict a model's performance in MT evaluation tasks. This finding suggests the need for better targeted intrinsic evaluation approaches that reflect downstream application requirements (\Cref{fig:heatmap}).
    \item \textbf{Temporal Stability:} The stability of the correlations over time between the segment-level metrics provides encouraging evidence for the reliability of these evaluation approaches.
    % TODO: mam pocit, ze ten predchozi odstavec uplne nedava smysl - promyslet si
    \item \textbf{Language-Specific vs. Multilingual Models:} There are no strong differences in performance between language-specific and multilingual models. Both categories are comparably represented among the top-performing models in intrinsic and extrinsic tasks.
    \item \textbf{Model Size Might Matter:}  In contrast to intrinsic tasks, fine-tuning embeddings for MTE/QE reveals that model size does matter, with the \emph{small} embeddings consistently showing poor performance.
   % \item ?Zabrusovat do tohoto? Model architecture  seems more important than whether it's multilingual or monolingual
\end{itemize}

In future work, we intend to replicate these experiments across multiple languages to investigate whether the observed behavior is specific to the Czech language or if it generalizes to other languages. In addition, we plan to conduct a more thorough analysis to better understand the underlying reasons for the differences in performance between the evaluation methods.

%These insights underscore the importance of aligning evaluation methodologies with downstream tasks and suggest that further research should explore refined intrinsic measures as well as the impact of architectural and size-related factors on model performance.

\section*{Acknowledgments}

The authors acknowledge the support of
the National Recovery Plan funded project MPO 60273/24/21300/21000 CEDMO 2.0 NPO
  % Petra, cestovne a registrace
and the project OP JAK Mezisektorová spolupráce Nr.\ CZ.02.01.01/00/23\_020/0008518 named ``Jazykověda, umělá inteligence a jazykové a řečové technologie: od výzkumu k aplikacím.''
  % Ondrej.
This work has been conducted using data provided by the LINDAT/CLARIAH-CZ Research Infrastructure (https://lindat.cz), supported by the Ministry of Education, Youth and Sports of the Czech Republic (Project No. LM2023062). 

% Bibliography entries for the entire Anthology, followed by custom (mtsummit25) entries
%\bibliography{anthology,mtsummit25}
\bibliography{paper}

%\appendix

%\section{Example Appendix}
%\label{sec:appendix}

%This is an appendix.

\end{document}